\documentclass[twocolumn,10pt]{asme2ej}

\usepackage{graphicx} 
\usepackage[english]{babel}
\usepackage{amsmath}
\usepackage[mathscr]{euscript}
\usepackage{mathrsfs}
\usepackage{times}
\usepackage[usenames,dvipsnames,svgnames,table, xcdraw]{xcolor} 
\usepackage{amsfonts}
\usepackage{url}
\usepackage{subcaption}
\usepackage{booktabs}
\usepackage{multirow}
\usepackage[ruled,vlined,linesnumbered]{algorithm2e}
\SetKw{Continue}{continue}
\SetKw{Break}{break}

\usepackage{enumitem}

\usepackage{cite} 

\newcommand{\eqcolor}[1]{{{\textcolor{blue}{#1}}}}
\newcommand{\modeldetheuristic}{D-LNS}
\newcommand{\modeldetexact}{D-ES}
\newcommand{\modeluncertain}{S-ES}

\usepackage[breaklinks=true,letterpaper=true,colorlinks,bookmarks=false]{hyperref}

\title{
Simultaneous Human-robot Matching and Routing for Multi-robot Tour Guiding under Time Uncertainty
}

\title{
Simultaneous Human-robot Matching and Routing for Multi-robot Tour Guiding under Time Uncertainty
}

\author{Bo Fu
    \affiliation{
	Robotics Institute\\
	University of Michigan\\
	Ann Arbor, MI 48109\\
    bofu@umich.edu
    }	
}

\author{Tribhi Kathuria
    \affiliation{
	Robotics Institute\\
	University of Michigan\\
	Ann Arbor, MI 48109\\
    tribhi@umich.edu
    }	
}

\author{Denise Rizzo
    \affiliation{
	DEVCOM GVSC\\
	US Army\\
	Warren, MI 48397\\
    denise.m.rizzo2.civ@army.mil
    }	
}

\author{Matthew Castanier
    \affiliation{
	DEVCOM GVSC\\
	US Army\\
	Warren, MI 48397\\
    matthew.p.castanier.civ@army.mil
    }	
}

\author{X. Jessie Yang
    \affiliation{
	Robotics Institute\\
	University of Michigan\\
	Ann Arbor, MI 48109\\
    xijyang@umich.edu
    }	
}

\author{Maani Ghaffari
    \affiliation{
	Robotics Institute\\
	University of Michigan\\
	Ann Arbor, MI 48109\\
    maanigj@umich.edu
    }	
}

\author{Kira Barton
    \affiliation{
	Robotics Institute\\
	University of Michigan\\
	Ann Arbor, MI 48109\\
    bartonkl@umich.edu
    }	
}

\renewcommand\footnotemark{}

\author{\thanks{Distribution A. Approved for public release; distribution unlimited. (OPSEC 5767)
}}

\begin{document}
\setlength\abovedisplayskip{5pt}
\setlength\belowdisplayskip{10pt}

\maketitle
\begin{abstract}
{\it 
This work presents a framework for multi-robot tour guidance in a partially known environment with uncertainty, such as a museum.
In the proposed centralized multi-robot planner, a simultaneous matching and routing problem (SMRP) is formulated to match the humans with robot guides according to their selected places of interest (POIs) and generate the routes and schedules for the robots according to uncertain spatial and time estimation.
A large neighborhood search algorithm is developed to efficiently find sub-optimal low-cost solutions for the SMRP.
The scalability and optimality of the multi-robot planner are evaluated computationally under different numbers of humans, robots, and POIs. The largest case tested involves 50 robots, 250 humans, and 50 POIs.
Then, a photo-realistic multi-robot simulation platform was developed based on Habitat-AI to verify the tour guiding performance in an uncertain indoor environment.
Results demonstrate that the proposed centralized tour planner is scalable, makes a smooth trade-off in the plans under different environmental constraints, and can lead to robust performance with inaccurate uncertainty estimations (within a certain margin).
}
\end{abstract}

\section{Introduction}\label{sec:introduction}
\begin{figure*}[t!]
	\centering
	\includegraphics[width=\linewidth]{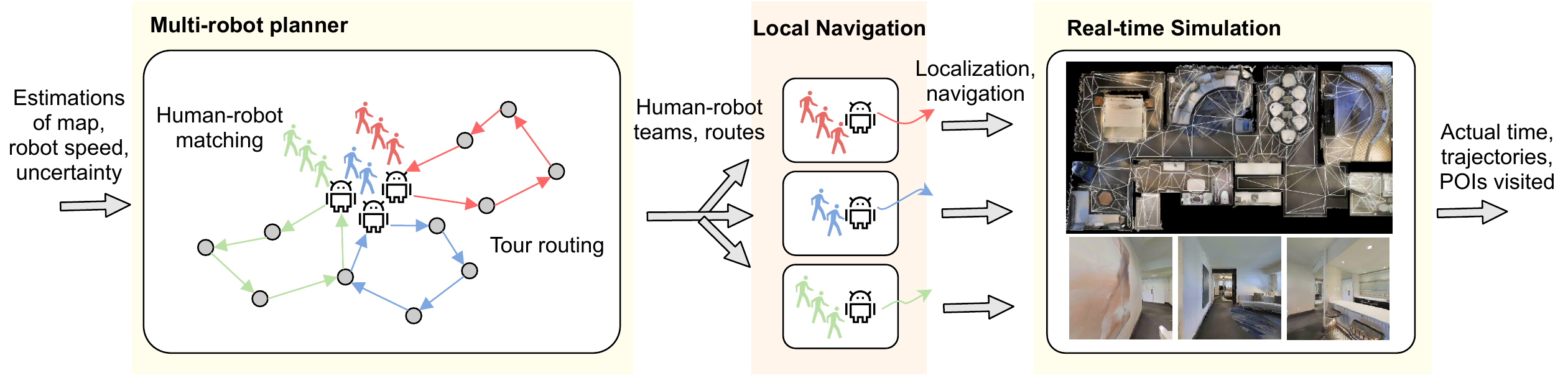}
	\caption{A framework for applying our multi-robot planner within a simulation. The centralized multi-robot tour planner is discussed in Sec. \ref{sec:global_planner}. The local navigation and simulation are described in Sec. \ref{sec:simulation_model}.}
	\label{fig:framework_diagram}
\end{figure*}

Robot tour guides have been applied in different environments \cite{satake2009approach, shiomi2009field, doering2019neural, kirby2010affective, burgard1998interactive, burgard1999museum} because they save human labor, do not require time-consuming training, and have the possibility to ease discomfort during social interactions \cite{de2019social}. Previous works have been focusing on the routing of places of interest (a traveling salesman style problem) \cite{preferences},
localization and navigation within the environment \cite{burgard1998interactive, burgard1999museum},
and social interactions with humans through gesture, face, and language processing \cite{bennewitz2005towards}.
While they cover diverse topics, these works mainly focus on single robot guiding for a small predetermined team.
In practice, it is unlikely that a single robot will be used to guide all of the human visitors; hence there is a need to use multiple robots to handle a larger number of humans. In such a situation, instead of considering each robot separately (as in previous works), a crucial question is how to coordinate the robots as a multi-robot system, such that the overall efficiency/performance of the system is maximized.

A natural first step is to consider loose coordination:
once the humans are split into teams and assigned to each robot,
the single-robot tour guide systems in previous works can be applied to the low-level navigation and interaction tasks.
However, the problems of choosing the best robot guides for a human and finding the optimal tour plan for a robot depend on each other.
The two problems are tightly coupled, leading to challenges to the optimization of both the tour plan and the human-robot matching.

Note that in this formulation, we consider homogeneous robots, and therefore, the local navigation and social interaction behaviors of the robots are homogeneous, do not affect, and therefore, are not considered during the multi-robot coordination at this time.
Future work may consider navigation and complex social interactions within the matching and touring problem.

This work focuses on a multi-robot planning system that simultaneously optimizes the human-robot team and a high-level robot tour plan specified as a route of places of interest (POIs).
The system provides an optimal matching between humans and robots and the corresponding tour plans.
The optimization planner proposed in this paper can be paired with a local navigation/interaction planner to form a complete framework that can be used to handle practical multi-robot guidance applications. Additionally, the touring system can be applied to different environments, such as a museum docent, a park tour, and a city tour.

As the multi-robot planning is performed premission, the actual touring times may vary due to dynamic movements of the teams, changes in the environment, and uncertain visiting times at a POI. Such variations will be considered as uncertainty in the traveling and visiting time in the optimization to generate time-robust touring plans.

A computational investigation is conducted with the proposed multi-robot planning system. Then, the system performance in an uncertain environment is evaluated through a photo-realistic simulation.
A system diagram of the simulation framework is shown in Fig. \ref{fig:framework_diagram}. A centralized multi-robot tour planner first generates the teams and tour plans for all robots. Then, individual robots use a local navigation planner to follow the planned routes and lead the humans within the simulation environment.

This work has the following contributions.
\begin{enumerate}[label={\arabic*)}]
    \item The algorithmic modeling of a \textbf{simultaneous} human-robot \textbf{matching} and tour \textbf{routing problem (SMRP)} for multi-robot tour guidance under execution time uncertainty.
    \item A comprehensive computational evaluation of the scalability and solution quality of the proposed algorithms.
    \item The simulation verification of the proposed multi-robot system through a concrete tour guiding case study in a photo-realistic indoor environment.
\end{enumerate}

The remainder of this paper is organized as follows:
Sec. \ref{sec:related_work} briefly introduces work related to socially assistive robots and the SMRP problem.
Sec. \ref{sec:global_planner} describes the math formulation and algorithms of the SMRP planner for multi-robot tour guides.
Sec. \ref{sec:simulation_model} introduces the local navigation planner and the simulation environment.
Note that sections \ref{sec:global_planner}-\ref{sec:simulation_model} also describe how uncertainties are considered in the planner and simulated in the simulation.
Sec. \ref{sec:experiment} provides computational and simulation results, followed by discussion.
Sec. \ref{sec:conclusion} concludes the paper.

\section{Related Work}\label{sec:related_work}

\begin{figure}[t!]
	\centering
	\includegraphics[width=0.37\linewidth]{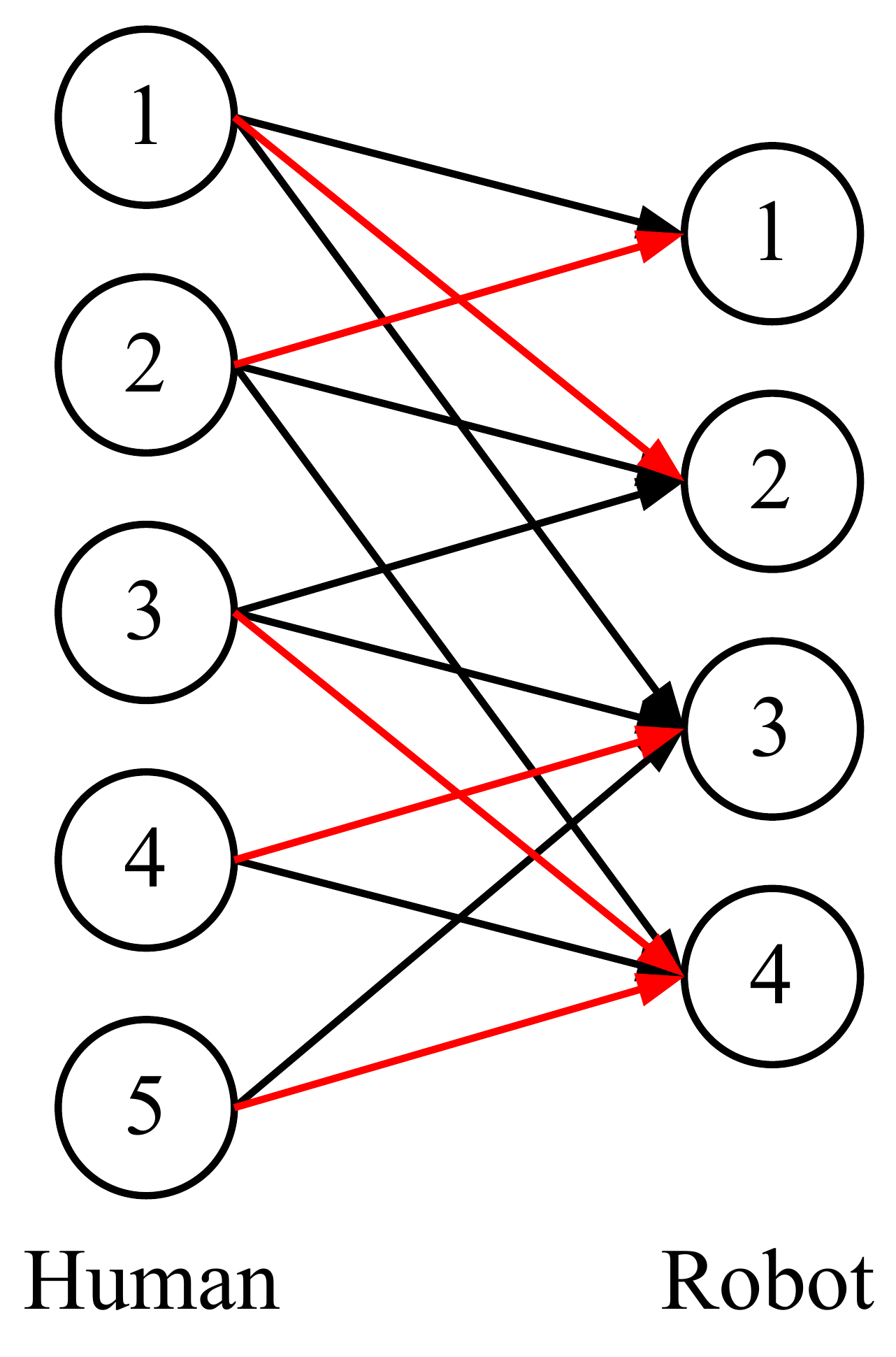}
	\caption{A bipartite matching example. A bipartite graph is a graph where there are two sets of nodes and no edge connections within each set. 
	A matching is a selected set of edges that satisfy some desired properties.
	In this paper, a matching should ensure that every human node is matched to one robot node (the red edges form a matching).
	The bipartite matching problem in the tour planning tries to find the best matching between humans and robots, such that the overall satisfied human requests are maximized.}
	\label{fig:matching_definition}
\end{figure}

\begin{figure}[t!]
	\centering
	\includegraphics[width=0.95\linewidth]{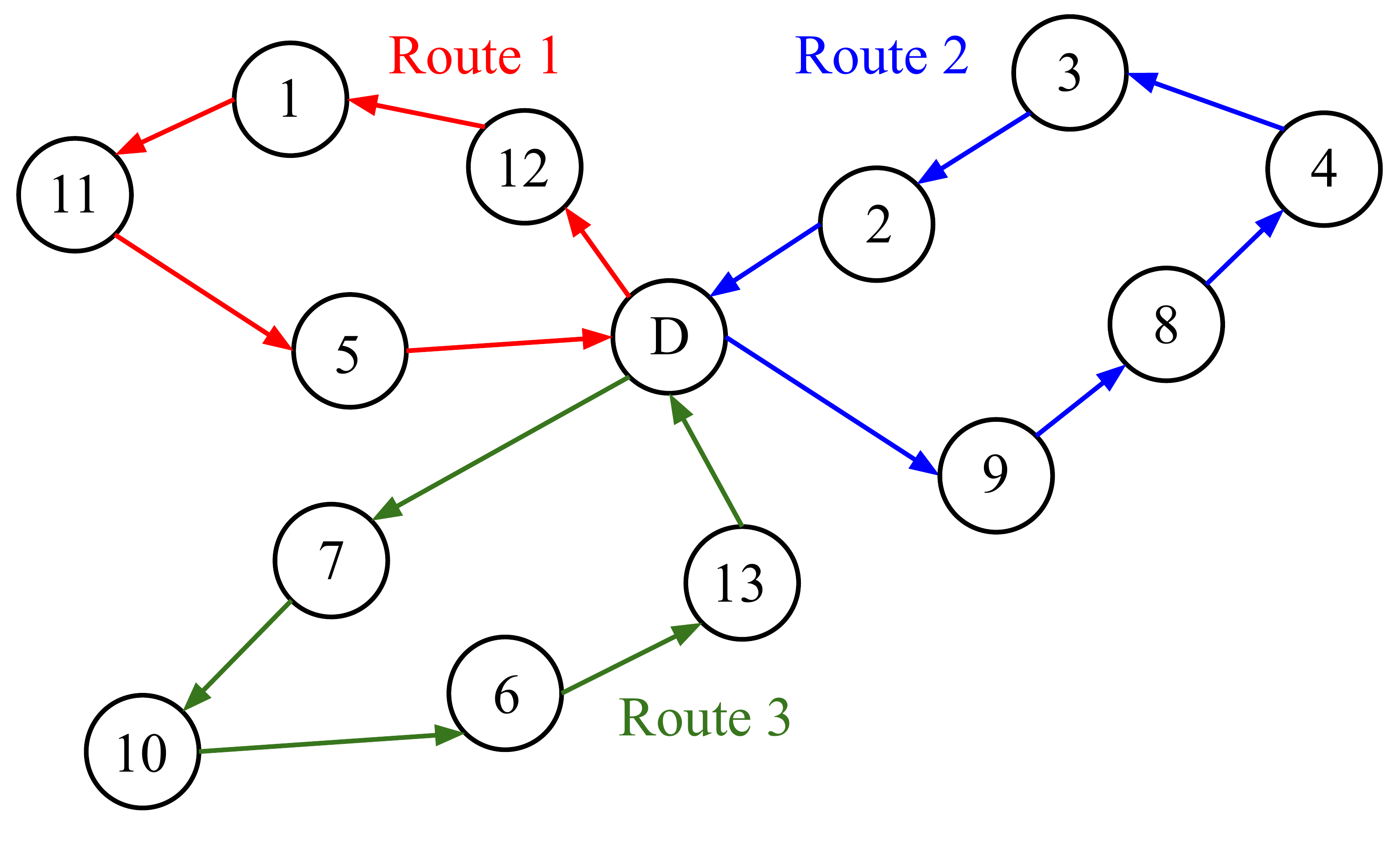}
	\caption{A vehicle routing problem (VRP) example with 3 vehicles. The VRP tries to find the minimum cost vehicle paths that satisfy some user-defined constraints.}
	\label{fig:routing_definition}
\end{figure}

In this work, the multi-robot tour guiding problem contains not only tour planning but also the human-robot team generation. Therefore, the optimization combines two classic problem types from the operations research: bipartite matching and vehicle routing. The matching problem refers to splitting the humans into teams such that the matching between humans and robots minimizes the number of dropped human requests. It is a bipartite matching problem as there are two separate sets of elements (robots and humans) in the problem. The definition of bipartite matching in the tour planning is illustrated in Fig. \ref{fig:matching_definition}. The routing problem focuses on optimally arranging the robot tours (routes), such that the dropped requests, as well as the traveling distance, are minimized (see Fig. \ref{fig:routing_definition}).

Optimal bipartite matching can be found through linear programming or maximum flow optimization \cite{ford2015flows}. Bipartite matching has been applied to multiple real-world problems including image feature matching \cite{cheng1996maximum}, object detection \cite{carion2020end}, budget allocation \cite{aggarwal2011online}, and task assignment \cite{dutta2019one}.

The vehicle routing problem (VRP) considers the minimum traversing distance of all the places of interest using multiple vehicles (robots). It generalizes the traveling salesman problem and is NP-hard to solve optimally. The stochastic vehicle routing problem (SVRP) is a variation of VRP where some of the parameters in the optimization are random distributions. Previous works in the field of SVRP have investigated optimization under uncertain requests number at each POI \cite{mendoza2013multi,secomandi2009reoptimization,marinakis2013particle, fu2021robust}, 
uncertain time costs for traveling edges or service at a POI \cite{sundar2017path,chen2014optimizing,li2010vehicle,gomez2016modeling},
and uncertain energy costs \cite{venkatachalam2019two,venkatachalam2018two, fu2020heterogeneous}. This work considers the uncertainty in time costs.

In this work, the optimal robot guide for a human depends on the robots' routes, while the route of a robot depends on the humans in its team. Therefore, the matching and routing problems are coupled to form a larger problem that must be optimized simultaneously. An SMRP problem contains multiple vehicle routing problems (formally, a VRP can be reduced to an SMRP); and since a VRP is NP-Hard \cite{toth2002vehicle}, an SMRP is NP-hard. A ride-sharing system can be regarded as an SMRP where the riders should be matched with drivers while the routes are determined simultaneously. However, many systems in previous work decouple the problems by generating the vehicle routes first and then matching humans with the closest routes \cite{schreieck2016matching, aydin2020matching}. Another way to formulate the ride-sharing problem is as a VRP with pickup and delivery \cite{sitek2019capacitated}.
Optimally solving the simultaneous matching and routing optimization under uncertainty can be challenging as this is a discrete non-convex problem that is NP-hard.
Unlike previous work, we directly address this simultaneous optimization problem, model it as a mixed-integer optimization, and provide methods that efficiently find sub-optimal low-cost solutions in Sec. \ref{sec:global_planner}.

\section{Human-robot Matching and Tour Routing}\label{sec:global_planner}

Consider the following practical situation: several human visitors arrive and each selects several places of interest. The guiding system wants to use a number of robots to guide these humans around the environment while satisfying as many POIs as possible. It is assumed that there are fewer robots than humans, and therefore, the people need to be split into teams. Since there is a time limit for the tours, not all human requests can be satisfied. There is uncertainty in the traveling and visiting time during the tour. A multi-robot planning system needs to find an optimal way to form the human-robot teams and the corresponding tours (defined as routes and schedules for the teams) to minimize the dropped requests under the aforementioned constraints and time uncertainties.

In this section, we first formally define the simultaneous matching and routing problem for the above practical multi-robot tour guidance. Then, we encode the problem as a mixed-integer program and provide a heuristic algorithm to solve the problem.

\subsection{Problem Description and Graphical Model}

Suppose there is a set of guidance robots \(V = \{1, \cdots, n_V\}\), a set of humans \(L = \{1, \cdots, n_L\}\), and a set of POIs to visit \(M = \{1, \cdots, n_M\}\). Each human can choose a subset of POIs that they want to visit. In this problem, a planner needs to determine which robot a human should follow and what routes (a sequence of POIs) a robot should take, such that the number of satisfied human requests (POIs) is maximized within certain touring time limits.

Other user-defined practical constraints include time window constraints (certain POIs are available only during fixed time windows), sequence dependencies (the visit of certain POIs has prerequisites that other POIs be visited first), and human dependencies (some people may prefer to be assigned to the same robots, such as families and friends).

\begin{figure}[t!]
	\centering
	\includegraphics[width=0.53\linewidth]{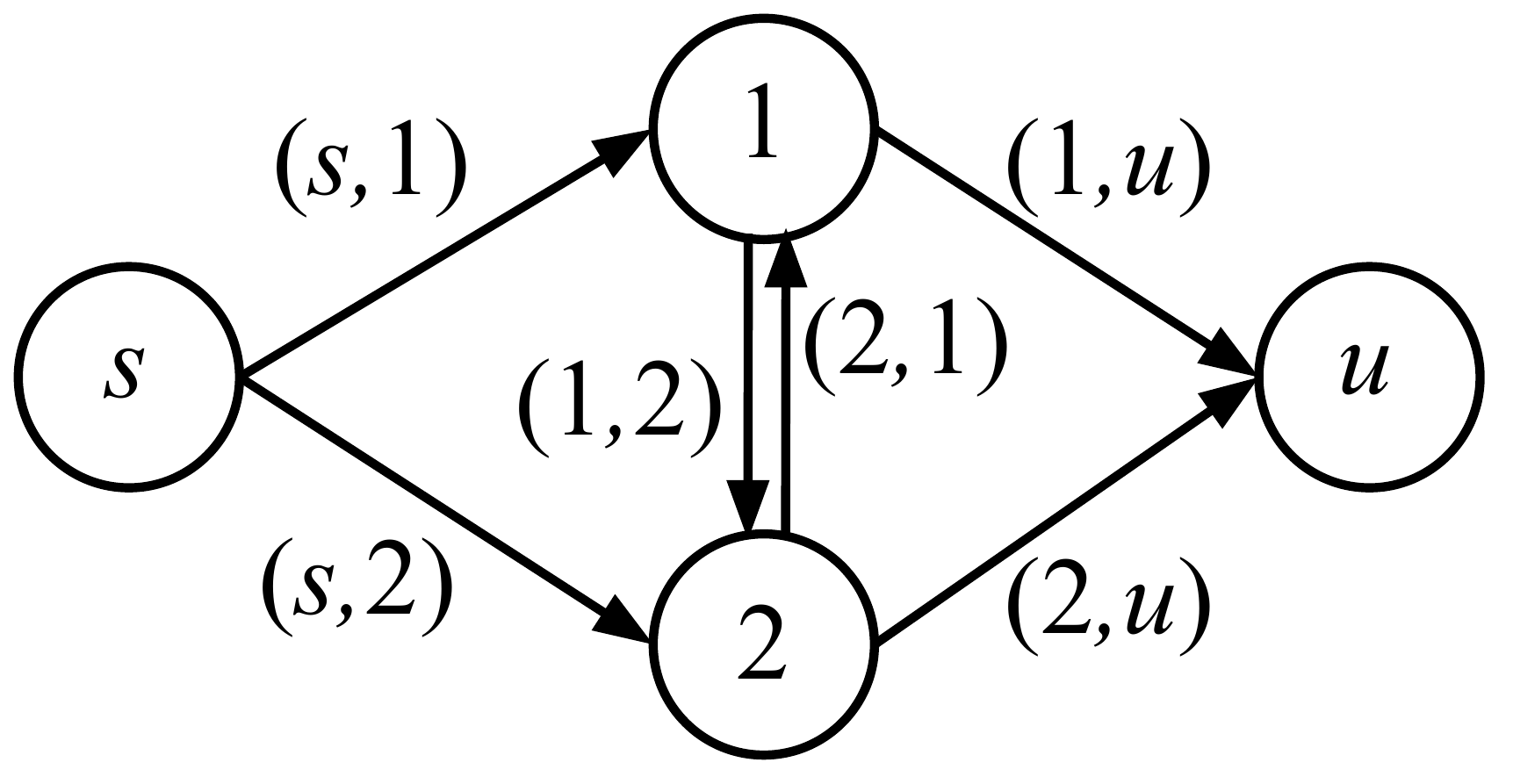}
	\caption{A graphical model example with two POIs. \(s\) and \(u\) are the start and terminal, respectively.}
	\label{fig:graphical_model}
\end{figure}

Suppose the start and terminal locations of the robots are nodes (places) \(n_M + 1\) and \(n_M + 2\). Namely, we have two additional nodes \(s = n_M + 1\) and \(u = n_M + 2\) in the graph. Then, let \(N = \{1, \cdots, n_M, s, u\}\) be the whole set of places. Note that, \(M\) is a subset of \(N\), and \(M\) does not include the start and terminal. We first define a directed graph \(G = (N, E)\), with \(N\) and \(E\) the sets of nodes and edges, respectively (see Figure \ref{fig:graphical_model}). For each node pair \(\{i,j\}\) with \(i, j \in N\), we add two directional edges \((i,j)\) and \((j,i)\) between them. Therefore, we have the edge set denoted as \(E = \{(i,j)\}, \ \forall i,j \in N\). Note that in our model, there is no assumption that the POIs are on a flat plane. If the tour happens in a multi-level building, no further modeling is needed as long as the edge costs are correctly adjusted according to the levels.

\subsection{A Mixed-integer Bilinear Program}\label{sec:deterministic_math}

\begin{figure}[t!]
	\centering
	\includegraphics[width=\linewidth]{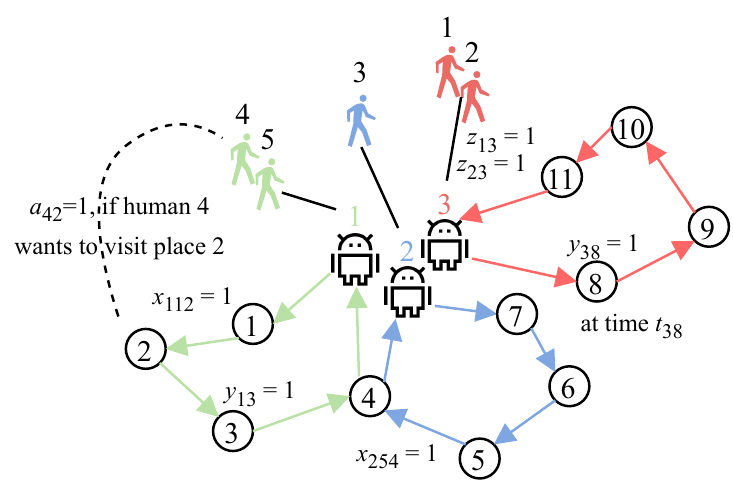}
	\caption{An illustration of the key notations through an example matching and routing.}
	\label{fig:framework_model}
\end{figure}

Here we discuss a mixed-integer bilinear program that mathematically models the SMRP we discussed above. We first provide the notations that will be used in Table \ref{tab:variable_definition}. Note that the decision variables are \(x_{kij}\), \(y_{ki}\), \(t_{ki}\), and \(z_{lk}\). Other symbols in Table \ref{tab:variable_definition} are predefined hyper-parameters or parameters that describe the environment and human-related information. A graphical illustration of the key notations is shown in Figure \ref{fig:framework_model}.

\begin{table}[h]
  \caption{Definition of the notations.}
  \label{tab:variable_definition}
    \small
    \begin{tabular}{p{0.06\linewidth}|p{0.8\linewidth}} 
    \toprule
     & Meaning
    \\
    \midrule
    \(x_{kij}\) & = 1, if robot \(k \in V\) travels edge \(i,j \in N\), otherwise 0. \\
    \(y_{ki}\) & = 1, if robot \(k \in V\) visits node \(i \in M\), otherwise 0. \\
    \(t_{ki}\) & The time robot \(k \in V\) visits node \(i \in N\). \\
    \(z_{lk}\) & = 1, if human \(l \in L\) is assigned to follow robot \(k \in V\), otherwise 0. \\
    \(a_{li}\) & = 1, if human \(l \in L\) wants to visit node \(i \in M\), otherwise 0. \\
    \(V\) & The set of robots. \\
    \(L\) & The set of humans. \\
    \(M\) & The set of POIs. \\
    \(N\) & The set of POIs plus the start and terminal. \\
    \(M_T\) & The set of POIs with additional time window constraints. \\
    \(S\) & The set of sequence dependencies. \\
    \(L_\mathrm{pair}\) & The set of human pairs who want to be in the same team. \\
    \(T_{L}\) & A large constant time. \\
    \(T_{k i}\) & The time for robot \(k\) and its team to visit POI \(i\). \\
    \(T_{k i j}\) & The time for robot \(k\) and its team to travel edge \((i,j)\). \\
    \(T_k^\mathrm{lim}\) & A preset time limit for the tour guided by robot \(k\). \\
    \(T_i^{\min}\) & The lower bound for a preset time window for visiting POI \(i\). \\
    \(T_i^{\max}\) & The upper bound for a preset time window for visiting POI \(i\). \\
    \(Z_k^{\max}\) & The maximum number of humans guided by robot \(k\). \\
    \(A^{\max}\) & The maximum number of POIs that can be dropped for one human. \\
    \bottomrule
    \end{tabular}
\end{table}

\noindent\textbf{Objective Function:}
In the objective function \eqref{eqn:bilinear_objective}, we want to minimize the weighted combination of dropped human requests and time usage, with the weights \(C_a\) and \(C_t\), respectively. The right part is the sum of the time that each human-robot team reaches the terminal node. The left part is the total number of dropped POIs for all humans. As an example, if human \(l\) follows robot \(k\), \(z_{l k}\) equals 1. If this human \(l\) wants to see POI \(i\), \(a_{l i}\) equals 1. In addition, if robot \(k\) does not pick POI \(i\) in its route, \(y_{k i}\) equals 0.
Finally, if \(a_{l i} = 1\), \(z_{l k} = 1\), and \(y_{k i} = 0\) happen simultaneously, the left part of the objective function will be incremented by 1, penalizing the drop of a human request. In this example, the weights are chosen such that \(C_a \gg C_t\) to ensure that the main focus is on satisfying human requests.
\begin{align}
    \min \
    C_a \cdot \sum_{l \in L} \sum_{k \in V} z_{l k}\left(\sum_{i \in M} a_{l i} \cdot\left(1-y_{k i}\right)\right) + C_t \sum_{k \in V} t_{k u}. \label{eqn:bilinear_objective}
\end{align}

\noindent\textbf{Variable Bounds:}
The feasible regions and bounds of four sets of decision variables are defined as in \eqref{eqn:var_bounds}.
\begin{align}
    x_{k i j}, \ y_{k i}, \ z_{l k} \in \{0,1\}, \quad t_{k i} \geq 0, \nonumber \\
    \forall k \in V, \ \forall i, j \in N, \ \forall l \in L. \label{eqn:var_bounds}
\end{align}

\noindent\textbf{Network Flow Constraints:}
Equation \eqref{eqn:flow_constraints1} is a network flow constraint that ensures the incoming robot number equals the outgoing robot number. Constraint \eqref{eqn:flow_constraints2} ensures that there is only one path in the network flow of robot \(k\). Constraints \eqref{eqn:var_relation_constraint1}-\eqref{eqn:var_relation_constraint2} show the relationship between variable \(x\) and \(y\). The number of robots at a node equals the number of incoming robots through the corresponding edges. Moreover, the number of robots at any node should be no larger than the number departing from the start node.
\begin{align}
    \sum_{i \in N} x_{k i m}=\sum_{j \in N} x_{k m j}, \quad & \forall m \in M, \quad \forall k \in V. \label{eqn:flow_constraints1} \\
    \sum_{i \in N} x_{k s i} \leq 1, \quad \quad \quad & \forall k \in V.  \label{eqn:flow_constraints2} \\
    y_{k j} \leq \sum_{i \in N} x_{k s i}, \quad \quad & \forall j \in M, \quad \forall k \in V.  \label{eqn:var_relation_constraint1} \\
    y_{k j} = \sum_{i \in N} x_{k i j}, \quad \quad & \forall j \in M, \quad \forall k \in V. \label{eqn:var_relation_constraint2}
\end{align}

\noindent\textbf{Time Constraints:}
Inequalities \eqref{eqn:time_constraint1}-\eqref{eqn:time_constraint2} are cumulative time constraints: if robot \(k\) travels edge \((i,j)\), then \(x_{k i j}=1\), and the difference between the arrival time at node \(j\) and node \(i\) equals the total time for visiting node \(i\) and traveling edge \((i,j)\) (i.e., \(T_{k i} + T_{k i j}\)). Constraint \eqref{eqn:time_limit_constraint} sets a time limit \(T_k^\mathrm{lim}\) for the whole tour. Constraint \eqref{eqn:time_window_constraint} sets a specific time window \([T_i^{\min}, \ T_i^{\max}]\) for visiting a node. This is important for some of the POIs that have a specific opening time (e.g., a film or a show). We use the set \(M_T \subset M\) to denote these specific POIs.
\begin{align}
    t_{k i} - t_{k j} + T_{k i j} + T_{k i} \leq T_{L} (1 - x_{k i j}), \ \ & \forall i, j \in N, \forall k \in V. \label{eqn:time_constraint1} \\
    t_{k i} - t_{k j} + T_{k i j} + T_{k i} \geq - T_{L} (1 - x_{k i j}), \ \ & \forall i, j \in N, \forall k \in V. \label{eqn:time_constraint2} \\
    t_{k u} \leq T_k^\mathrm{lim}, \quad \quad \quad \quad &\forall k \in V. \label{eqn:time_limit_constraint} \\
    T_i^{\min} \leq t_{k i} \leq T_i^{\max}, \quad \quad &\forall k \in V, \forall i \in M_{T}. \hspace{-0.1cm} \label{eqn:time_window_constraint}
\end{align}

\noindent\textbf{Sequence Dependency Constraints:}
There can be sequence dependencies between the POIs. For example, if a person does not visit POI \(i\), they would not be able to gain the knowledge needed to understand and appreciate the work at POI \(j\). Let \(S\) denote the set of the sequence dependencies. As an example, \(S\) can be denoted as \(\{(1,2), (2,3), (6,8)\}\), which means that POI 2 depends on 1, and POI 3 depends on 1 and 2, while POI 8 has POI 6 as a prerequisite. Mathematically, the sequence dependency is encoded as in \eqref{eqn:artistic_dependency}. The first inequality ensures that node \(i\) has to be visited if a route visits node \(j\). The second inequality related to time ensures that node \(i\) is visited before \(j\).
\begin{align}
    y_{k i} \geq y_{k j}, \ \ \ t_{k i} \leq t_{k j} + T_L (1 - y_{k j}), \ \
    \forall k \in V, \ \forall (i, j) \in S. \hspace{-0.1cm} \label{eqn:artistic_dependency}
\end{align}

\noindent\textbf{Team Size Limit Constraints:}
All humans must be matched to a robot guide, introducing the constraint \eqref{eqn:visitor_assignment_constraint}.
We also impose a limit to the maximum number of humans that follow a robot to avoid imbalanced teams as in \eqref{eqn:team_size_constraint}. This limit is denoted as \(Z_k^{\max}\). Due to this constraint, a human might not be matched to their optimal robot guide for balancing the assignment and minimizing overall dropped POIs. However, we add constraint \eqref{eqn:max_drop_constraint} to limit the maximum number of dropped requests, \(A^{\max}\), to avoid extreme sacrifices of specific humans' needs.
\begin{align}
    \sum_{k \in V} z_{l k} = 1, \quad \quad \quad \quad & \forall l \in L. \label{eqn:visitor_assignment_constraint} \\
    \sum_{l \in L} z_{l k} \leq Z_k^{\max}, \quad \quad \quad \quad & \forall k \in V. \label{eqn:team_size_constraint} \\
    \sum_{k \in V} z_{l k} (\sum_{i \in M} a_{l i} \cdot\left(1-y_{k i}\right)) \leq A^{\max}, \quad & \forall l \in L. \label{eqn:max_drop_constraint}
\end{align}

\noindent\textbf{Human Dependency Constraints:}
There can be human dependencies where some people want to be guided by the same robot guide (e.g., families and friends). Such a dependency is encoded in \eqref{eqn:visitor_dependency}, which forces two people to be in the same team. \(L_\mathrm{pair}\) is the set of human pairs that have such dependencies. An example can be \(L_\mathrm{pair} = \{(1,2), (2,3), (5,6)\}\).
\begin{align}
    z_{l_1 k} = z_{l_2 k}, \quad \forall k \in V, \ \ \forall (l_1, l_2) \in L_\mathrm{pair}. \label{eqn:visitor_dependency}
\end{align}

\subsection{Uncertain Travel and Visiting Time} \label{sec:stochastic_math}
In the above formulation, we do not consider time uncertainty. When there is uncertainty in the traveling time \(T_{kij}\) and visiting time \(T_{ki}\), we still need to make sure that the tour ends on time, i.e., constraint \eqref{eqn:time_limit_constraint} still holds. In such situations, \(T_{kij}\), \(T_{ki}\), and \(t_{k u}\) can be modeled as random distributions.
And we penalize the scenarios where the finishing time \(t_{k u}\) is close to the time limit \(T_k^\mathrm{lim}\) by adding an expected penalty, \(\phi_(t_{k u})\),
\begin{align}
    \phi_(t_{k u}) =& \mathrm{E}\left([t_{k u} - \alpha_k]^+\right)=\lim_{n_\xi \rightarrow \infty} \frac{1}{n_\xi}\sum_{\xi=1}^{n_\xi}\left[t_{ku}^\xi - \alpha_k\right]^+, \nonumber \\ 
    \alpha_k =& T_k^\mathrm{lim} - \Delta T_k^\mathrm{lim}, \nonumber \\
    [x]^+ =& \begin{cases} x, \quad x > 0 \\
    0, \quad x \leq 0 
    \end{cases}. \nonumber
\end{align}

\(\alpha_k\) is a time threshold when the penalty starts being applied and \(\Delta T_k^\mathrm{lim}\) is a time margin which is usually chosen as a small number (e.g., 5 minutes). \(t_{ku}^\xi\) is a sample of the distribution \(t_{k u}\) with \(\xi = 1, \cdots, n_\xi\). Such a penalty model has shown effectiveness in previous work in VRP with stochastic time cost \cite{sundar2017path}.
In practice, we only calculate an approximation of the \(\phi_(t_{k u})\) with a bounded number of samples as follows
\begin{align}
    \hat{\phi}(t_{ku}) = \frac{1}{n_\xi}\sum_{\xi=1}^{n_\xi}\left[t_{ku}^\xi - \alpha_k\right]^+. \nonumber 
\end{align}

Based on the derivation above, when there is time uncertainty, we modify the objective in \eqref{eqn:bilinear_objective} and minimize the new objective function \eqref{eqn:linearized_uncertain_objective1}, subject to the original constraints.
\begin{align}
    \min \ &
    C_a \cdot \sum_{l \in L} \sum_{k \in V} z_{l k} (\sum_{i \in M} a_{l i} \cdot\left(1-y_{k i}\right) ) + C_t \sum_{k \in V} \hat{\phi} \left( t_{k u} \right) \nonumber \\ 
    = \
    & C_a \cdot \sum_{l \in L} \sum_{k \in V} z_{l k} (\sum_{i \in M} a_{l i} \cdot\left(1-y_{k i}\right) ) \label{eqn:linearized_uncertain_objective1} \\
    & + C_t \sum_{k \in V} \frac{1}{n_\xi} \sum_{\xi=1}^{n_\xi}  \left[ t_{k u}^\xi - \alpha_k \right]^+. \nonumber
\end{align}

In addition, the total tour time \(t_{k u}\) equals the sum of all traveling time and visiting time as in
\begin{align}
    t_{k u} = \sum_{i \in N} \sum_{j \in N} T_{kij} \cdot x_{kij} + \sum_{i \in N} T_{ki} \cdot x_{ki}, \quad \forall k \in V. \label{eqn:sum_tour_time}
\end{align}

Suppose we know and can represent the random distributions of \(T_{kij}\) and \(T_{ki}\) as a series of samples \(T_{kij}^\xi\) and \(T_{ki}^\xi\) (\(\xi = 1, \cdots, n_\xi\)). Then, substituting \eqref{eqn:sum_tour_time} into \eqref{eqn:linearized_uncertain_objective1}, and linearizing the function \([x]^+\) using inequalities \eqref{eqn:w_constraint1}-\eqref{eqn:w_constraint2}, the original objective function \eqref{eqn:bilinear_objective} is replaced with \eqref{eqn:linearized_uncertain_objective2}. Therefore, in the situation with time uncertainty, the optimization problem becomes the following
\begin{align}
    \min \ &
    C_a \sum_{l \in L} \sum_{k \in V} z_{l k} (\sum_{i \in M} a_{l i} \cdot\left(1-y_{k i}\right) ) 
    + C_t \sum_{k \in V} \frac{1}{n_\xi} \sum_{\xi=1}^{n_\xi} w_k^\xi \label{eqn:linearized_uncertain_objective2}  \\
    \text{sub } & \text{to } \eqref{eqn:var_bounds}-\eqref{eqn:visitor_dependency} \text{ and } \nonumber \\
    w_k^\xi \geq & \sum_{i \in N} \sum_{j \in N} T_{kij} \cdot x_{kij} + \sum_{i \in N} T_{ki} \cdot x_{ki} - \alpha_k, \ \ \ \forall k \in V, \ \forall \xi, \label{eqn:w_constraint1} \\
    w_k^\xi \geq & 0, \quad \quad \quad \quad \quad \quad \quad \quad \quad \quad \quad \quad \quad \ \ \ \forall k \in V, \ \forall \xi.  \label{eqn:w_constraint2}
\end{align}

Note that \(w_{k}^\xi\) is a helper variable.

\subsection{A Branch-and-Cut Approach for an Exact Solution}

The optimization problem (with or without uncertainty) described in the above section is a mixed-integer quadratic program (MIQP) with bilinear objective function and linear constraints. Such a MIQP can be solved through a branch-and-cut algorithm where multiple continuous quadratic optimization problems are solved iteratively. A branch-and-cut algorithm tries to find an exact solution for an MIQP: it gives a solution and the optimality gap associated with the solution that indicates how close the solution is to the global optimum. Given enough time, such an exact algorithm will ultimately find the global optimum for a problem. Some available commercial solvers implement branch-and-cut algorithms. We encode the presented math formulation using the GUROBI solver.

\subsection{A Large Neighborhood Search Approach}\label{sec:large_neiborhood_search}

Exact algorithms are guaranteed to find the global optimum given enough time. However, for an NP-hard problem like the SMRP in this paper, the exact optimum requires exponentially growing computations. Therefore, exact solution algorithms are usually applied to problems of smaller sizes. Here, we provide a heuristic algorithm that provides potential solutions with low objective values for larger problem cases.

According to the formulation in Sec. \ref{sec:deterministic_math}, the only nonlinear part in the optimization problem is the bilinear term in the objective function \eqref{eqn:bilinear_objective}, if we fix either of the variables \(z_{l k}\) or \(y_{k i}\), and optimize the sub-problem with the remaining variables, the optimization becomes linear. Therefore, to solve the optimization, we can fix part of the variables, limit the search space to a subset of the whole feasible domain, find the solution for a simpler linear sub-problem, and then change the search space and repeat the process to improve the solution.

The idea of solving a larger problem by iteratively solving still-large but simpler sub-problems is called a large neighborhood search (LNS). The difference between LNS and a small neighborhood search (SNS), such as the gradient descent algorithm and trust-region algorithms, is that in each iteration, the size of the search space is much larger, which ends up with larger step sizes and fewer iterations. The LNS algorithm has been widely adopted and has shown its efficiency in many scheduling and routing-related applications \cite{chen2014optimizing, he2018improved, alinaghian2018multi, deng2021room}.

We now describe how we apply an LNS algorithm to solve the SMRP problem with and without time uncertainty in Sec. \ref{sec:deterministic_math} and \ref{sec:stochastic_math}.
The pseudo-code of the large neighborhood search algorithm for SMRP is in Algorithm \ref{alg:large_neighborhood_search}. The input of the algorithm is the unsolved optimization problem with all predefined parameters. From a set of randomly initialized variable values, the algorithm optimizes a matching sub-problem and \(n_V\) (the number of robots) routing sub-problems in each step to improve the solution quality iteratively. Note that both sub-problems are linear and with fewer variables, and therefore, much easier to solve than the original SMRP. Since the solving processes of the two sub-problems interleave in each step, the quality of the matching and routing are simultaneously optimized. The following paragraphs describe the matching sub-problem and routing sub-problem in detail.

\begin{algorithm}[t]
\small

\textbf{Input:} the unsolved optimization problem in Sec. \ref{sec:deterministic_math}-\ref{sec:stochastic_math}

Randomly initialize \(x_{kij}\), \(y_{ki}\), \(z_{lk}\), \(t_{ki}\)

\For{\textnormal{iteration} \(= 1, \cdots, n_{\max}\)}{
    Fix  \(x_{kij}\), \(y_{ki}\), \(t_{ki}\) \(\quad (\forall k \in V, \ \ \forall i,j \in N)\)
    
    Solve the matching sub-problem in \eqref{eqn:matching_subproblem}
    
    Update \(z_{lk}\)  \(\quad (\forall k \in V, \ \ \forall l \in L)\)
    
    \For{\(k \in V\)}{
        Fix \(z_{lk}\) \(\quad (\forall l \in L)\)
        
        Solve the routing sub-problem in \eqref{eqn:routing_subproblem1} or \eqref{eqn:routing_subproblem2}
        
        Update \(x_{kij}\), \(y_{ki}\), \(t_{ki}\) \(\quad (\forall i,j \in N)\)
    }
    
    \If{\textnormal{the objective value does not change}}
    {
        \Break
    }
}

\Return the solution \(x_{kij}\), \(y_{ki}\), \(z_{lk}\), \(t_{ki}\)

\caption{Large Neighborhood Search}
\label{alg:large_neighborhood_search}
\end{algorithm}

\textbf{Matching Sub-problem:}
Fix the variables \(x_{kij}, y_{ki}, t_{ki}\) and solve the SMRP optimization within the neighborhood of \(z_{lk}\). Both the problems with and without uncertainty reduce to the following optimization problem (variables are highlighted in blue).
\begin{align}
    \min & \ \
    \sum_{l \in L} \sum_{k \in V} \eqcolor{z_{l k}} (\sum_{i \in M} a_{l i} \cdot\left(1-y_{k i}\right) ) \label{eqn:matching_subproblem} \\
    \text{sub to} & \ \ \eqcolor{z_{lk}} \in \{0, 1\} \label{eqn:z_constraint} \\
    & \text{and} \ \ \eqref{eqn:team_size_constraint}-\eqref{eqn:visitor_dependency}. \nonumber
\end{align}

This is a smaller integer linear program (ILP) with fewer variables than the original SMRP and can be solved much faster. If there is no maximum dropped requests constraint \eqref{eqn:max_drop_constraint} or human dependency constraint \eqref{eqn:visitor_dependency}, then, this is a standard bipartite matching problem whose integer solutions can be obtained by solving the reduced linear program (removing constraint \eqref{eqn:z_constraint}) in a polynomial time \cite{heller1956extension}. We encode the math formulation here using GUROBI.

\textbf{Routing Sub-problem:}
Fix variable \(z_{lk}\) and solve the SMRP optimization within the neighborhood of \(x_{kij}, y_{ki}, t_{ki}\). Mathematically, the problem then reduces to a mixed-integer linear program (MILP). In terms of practical application, it becomes a variation of the vehicle routing problem.

Moreover, since the human-robot matching is fixed (fix variables \(z_{lk}\)), a robot only needs to consider the requests of the people in its own team. Therefore, the optimization is decoupled into multiple single-vehicle routing problems (i.e., traveling salesman problems). Mathematically, this means the optimization of \(x_{kij}, y_{ki}, t_{ki}\) decouples between different robots \(k \in V\). 
Using the large neighborhood search algorithm, we solve \(n_{V}\) single-vehicle routing sub-problems.

The decoupled routing problem for a single problem is shown below. For this version without uncertainty, we can leverage established routing solvers, which efficiently return low-cost solutions using multiple heuristics. Here we choose the Google Or-Tools to encode this model.
\begin{align}
    \min & \ \
    C_a \sum_{l \in L} z_{l k}\left(\sum_{i \in M} a_{l i} \cdot\left(1- \eqcolor{y_{k i}}\right)\right) + C_t \ \eqcolor{t_{k u}} \label{eqn:routing_subproblem1} \\ 
    \text{sub to} & \ \ \eqref{eqn:var_bounds}-\eqref{eqn:artistic_dependency}. \nonumber
\end{align}

The sub-problem formulation with uncertain traveling and visiting times is shown below. Again, we encode the math formulation here using GUROBI.
\begin{align}
    \min \ &
    C_a \sum_{l \in L} z_{l k} (\sum_{i \in M} a_{l i} \cdot\left(1- \eqcolor{y_{k i}}\right) )
    + \frac{C_t}{n_\xi} \sum_{\xi=1}^{n_\xi} \eqcolor{w_k^\xi} \label{eqn:routing_subproblem2} \\ 
    \text{sub} \  & \text{to} \ \ \eqref{eqn:var_bounds}-\eqref{eqn:artistic_dependency}, \ \text{and } \eqref{eqn:w_constraint1}-\eqref{eqn:w_constraint2}. \nonumber
\end{align}

Note that in \eqref{eqn:routing_subproblem1} and \eqref{eqn:routing_subproblem2}, the decision variables are \(y_{k i}\), \(t_{k u}\), and \(w_k^\xi\), while \(z_{l k}\) is fixed.

\section{Local Navigation and multi-robot Simulation}\label{sec:simulation_model}
In this section, we describe a simulation environment that we developed for testing the performance of the proposed framework under uncertainty. An indoor environment is used to demonstrate the possible use of the system in a museum environment. In addition, to complete the framework, we also discuss the local navigation planner and the strategy used. The simulation pipeline is implemented in ROS \cite{quigley2009ros}, which contains a multi-robot tour planner ROS node, multiple local planner nodes for a single robot, and a simulation wrapper node.

\subsection{Local Navigation Planner}

A local navigation planner generates the actions according to the current robot position and the routes generated by the multi-robot tour planner. Listed in Table \ref{tab:action_space}, there are four discrete actions that a robot can take. Since localization and mapping is not the focus of this research, robot positions are assumed to be known in the simulation.

\subsection{Habitat-AI Simulation Environment}

\begin{figure}
    \centering
    \includegraphics[width = \linewidth, trim=0 80 0 90, clip]{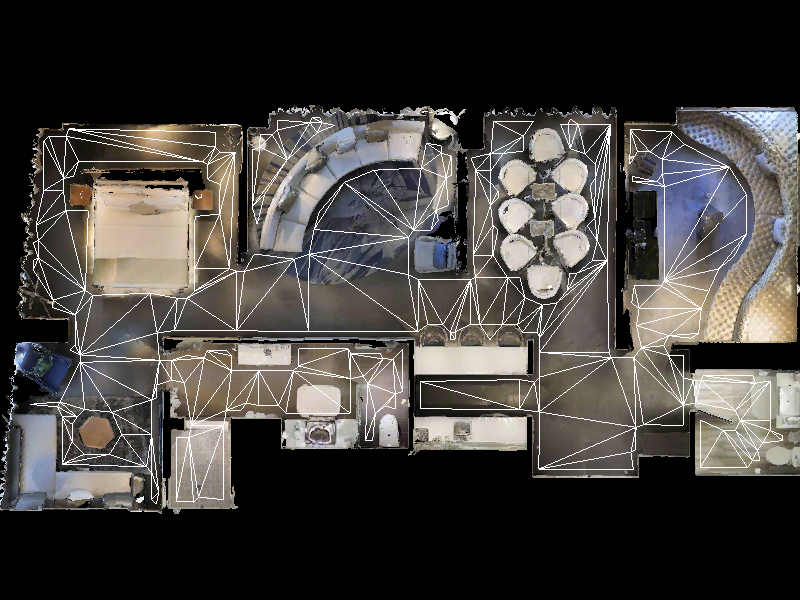}
    \caption{Top view of the chosen scene within the Matterport3D dataset to give a house tour within the Habitat-AI simulation environment. The white bounding boxes describe the navigable area in the scene, rendered as meshes.}
    \label{fig:house}
\end{figure}

\begin{figure}
\includegraphics[width=0.32\linewidth, trim=0 0 0 0, clip]{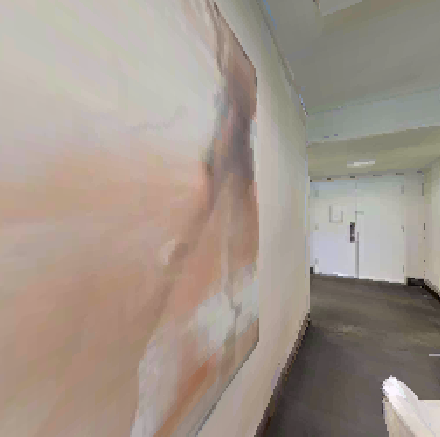}
\includegraphics[width=0.32\linewidth, trim=0 0 0 0, clip]{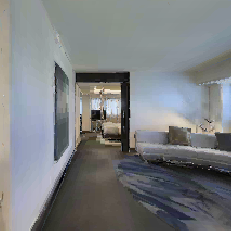}
\includegraphics[width=0.32\linewidth, trim=0 0 0 0, clip]{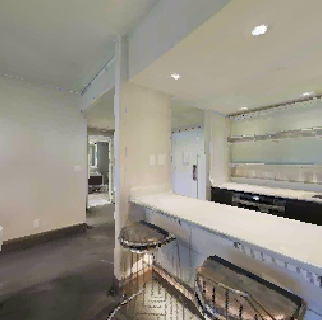}
	\caption{Example views of the robots conducting tasks.}
	\label{fig:camera_view}
\end{figure}

A real-time simulation environment is used to simulate the dynamics and obstacle avoidance behavior of the robots involved in this work.
The simulation is implemented in the Habitat-AI environment, a photo-realistic, physics-based simulator, to ensure that the framework and test results are transferable to real-world deployment \cite{savva2019habitat, chang2017matterport3d, kadian2020sim2real}
We choose a scene from the Matterport3D dataset, which is a collection of 90 scans, to design a sample tour of an indoor environment. Though the simulated environment is a house, it contains most of the features that would exist in a museum.
The top view of the chosen scene is depicted in Fig. \ref{fig:house}. And three first-person views of the robots are shown in Fig. \ref{fig:camera_view}.

\begin{table}[b]
    \caption{The discrete action space and its continuous action counterpart for the robot.}
    \label{tab:action_space}
    \centering
    \footnotesize
    \begin{tabular}{l|l}\toprule
        0 & STOP \\ \midrule
        1 & MOVE\_FORWARD (0.25 m) \\ \midrule
        2 & TURN\_LEFT ($10 ^{\circ}$) \\ \midrule
        3 & TURN\_RIGHT ($10 ^{\circ}$) \\ \bottomrule
    \end{tabular}
\end{table}

\subsection{Uncertainties in Navigation}

While the planned tour from the high-level planner specifies the path that a robot should execute, the actual paths of the robots have some inherent uncertainties. The tour planner considers such uncertainties as edge traveling time (between two POIs) and node visiting time (at each POI).
This section describes the factors in the simulation that introduce uncertainties to the traveling and visiting time.
\begin{figure}
    \centering
    \includegraphics[width=\linewidth]{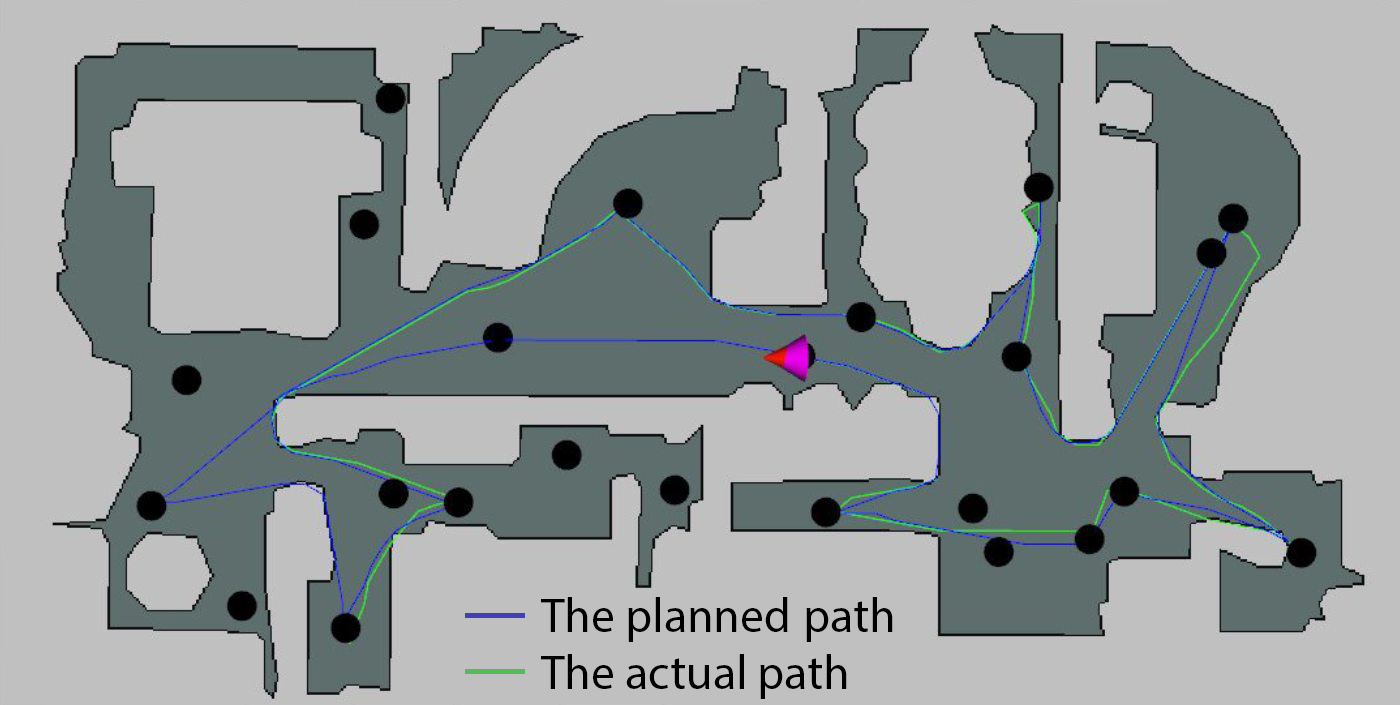}
    \caption{The planned path from the tour planner and the actual path in the presence of action uncertainty (10\% random actions). Due to the overlap, the actual path is only visible when the robot deviates from the ideal behaviors.}
    \label{fig:action}
\end{figure}
\subsubsection{Uncertainties in the Traveling Time}
\textbf{Disturbances and Imperfect Actions:}
Due to the lack of an accurate map and the dynamic objects in the world, the robot might need to adjust the planned tour routes according to the actual environment changes. A robot could also take wrong actions because of imperfect perception, action selection, and actuation. In the simulation, these factors are modeled as incorrect robot actions, which inject uncertainties to the edge time. The actions here are a set of movements (listed in Table \ref{tab:action_space}) that the robots can take. For instance, if the correct action for the robot is to move forward and it actually turns left, an incorrect action happens. We add a rate of correct robot actions in the simulator to control the level of incorrectness. As an example, when this rate is set to 90\%, a robot only takes the correct actions 90\% of the time and random actions for the remaining 10\%. The effect of 10\% incorrect actions is depicted in Fig. \ref{fig:action}.

\textbf{Robot Dynamics:}
Another set of uncertainties in edge time comes from the robot dynamics. The multi-robot tour planner estimates the traveling time according to traveling distance and average robot velocities. However, the actual dynamics of the robot system depend on the sequence of the actual robot actions and results in varying velocity and in-place rotations. The dynamics lead to variations of the actual traveling time, which is also considered a traveling time uncertainty. Note that this is an uncertainty that we are not able to control in the simulator.

\subsubsection{Uncertainties in the Visiting Time}

The interaction time is uncertain for a robot tour guide interacting with people at a POI.
This is considered as a visiting time uncertainty in the multi-robot tour planner, and robust plans are generated to ensure the satisfaction of the user-specified constraints of time under uncertainty.
In the simulator, the visiting time at each POI is sampled from a Gaussian distribution centered at a manually defined nominal time.

\section{Experiments and Results}\label{sec:experiment}
In this section, we first use randomly generated cases to evaluate the scalability and optimality of the proposed planning algorithms. These computational experiments are done on a laptop with an Apple M1 chip, and the corresponding results are shown in Sec. \ref{sec:computational_experiments}. Then we evaluated both the multi-robot tour planner in a photo-realistic simulation environment where the human's visiting time at a POI is uncertain and the robot could perform imperfect actions (which result in longer execution time). The simulation results are described in Sec. \ref{sec:simulation_experiments}.

\subsection{Computational Evaluation of the Tour Planning} \label{sec:computational_experiments}

We proposed two math models: a deterministic formulation and a stochastic formulation considering uncertain traveling and visiting time. We also propose two solution algorithms for the math models: an exact solution method that tries to find the global optimum and a LNS-based method that returns a heuristic solution. We will evaluate three of the combinations listed in Table \ref{tab:algorithm_tested}.

\begin{table}[tbp]
  \centering
  \small
  \caption{Algorithm types tested.}
    \begin{tabular}{c|l}
    \toprule
    \multicolumn{1}{c|}{Abbr} & \multicolumn{1}{c}{Meaning} \\
    \midrule
    \modeldetheuristic{} & Deterministic formulation with LNS algorithm \\
    \modeldetexact{}     & Deterministic formulation with exact solution method \\
    \modeluncertain{}    & Stochastic formulation with exact solution method \\
    \bottomrule
    \end{tabular}
  \label{tab:algorithm_tested}
\end{table}

\subsubsection{Scalability Evaluation}

The size of the problem mainly depends on the robot number \(n_V\), human number \(n_L\), and the total number of POIs in the area \(n_M\). Here we variate these three hyper-parameters and test the scalability of the algorithms under the randomly generated test cases. Specifically, We choose \(n_V \in \{4, 10, 20, 50\}\), \(n_L \in \{10, 50, 100, 250\}\), and \(n_M \in \{10, 20, 30, 50\}\). For other hyper-parameters, we choose a fixed value for all cases as they are not related to the scalability test. Particularly, for the scalability evaluation, the penalty for dropping a request, \(C_a\), is set to 1000, while the penalty on the time, \(C_t\), is set to 1, to ensure the dominant goal is to satisfy human requests. For each test case, we set a limit of 120 seconds for the solvers. For optimality, we use the dropped requests ratio, which is defined as (dropped requests / total requests) to indicate the solution quality. 

\begin{table}[t]
  \centering
  \small
  \caption{The dropped requests ratio of \modeldetheuristic{}. The \fbox{results in boxes} are where the result of D-LNS are larger (worse) than the results from the exact method \modeldetexact{}, results without boxes are the opposites.}
    \begin{tabular}{c|cccc}
    \toprule
          & \(n_V=\)4     & \(n_V=\)10    & \(n_V=\)20    & \(n_V=\)50 \\
          & \(n_L=\)10    & \(n_L=\)50    & \(n_L=\)100   & \(n_L=\)250 \\
    \midrule
    \(n_M=\)10    & \fbox{{0.06}}  & \fbox{{0.20}}  & \fbox{{0.26}}  & 0.27 \\
    \(n_M=\)20    & 0.00  & 0.26  & 0.31  & 0.45 \\
    \(n_M=\)30    & 0.00  & 0.20  & 0.31  & 0.44 \\
    \(n_M=\)50    & 0.03  & 0.12  & 0.28  & 0.36 \\
    \bottomrule
    \end{tabular}
  \label{tab:normalized_dropped_demand}
\end{table}

Results show that the largest problem where \modeldetexact{} and \modeluncertain{} can find a non-trivial solution within 120 seconds for a problem consisting of 10 robots, 50 humans, and 50 POIs (a trivial solution is defined as dropping all requests and conducting no tour). In contrast, \modeldetheuristic{} can return a non-trivial solution for all the cases. For the case with 50 robots, 250 humans, and 50 POIs, the \modeldetheuristic{} completes the optimization within 26.5 seconds and 7 iterations. The corresponding dropped requests ratio of \modeldetheuristic{} is listed in Table \ref{tab:normalized_dropped_demand}. According to the table, for the same deterministic SMRP, the LNS-based algorithm \modeldetheuristic{} outputs a better solution within 120 seconds except for the smallest cases.

\subsubsection{Time Limit and Dropped Requests}

\begin{figure}[h]
    \centering
    \includegraphics[width=0.9\linewidth]{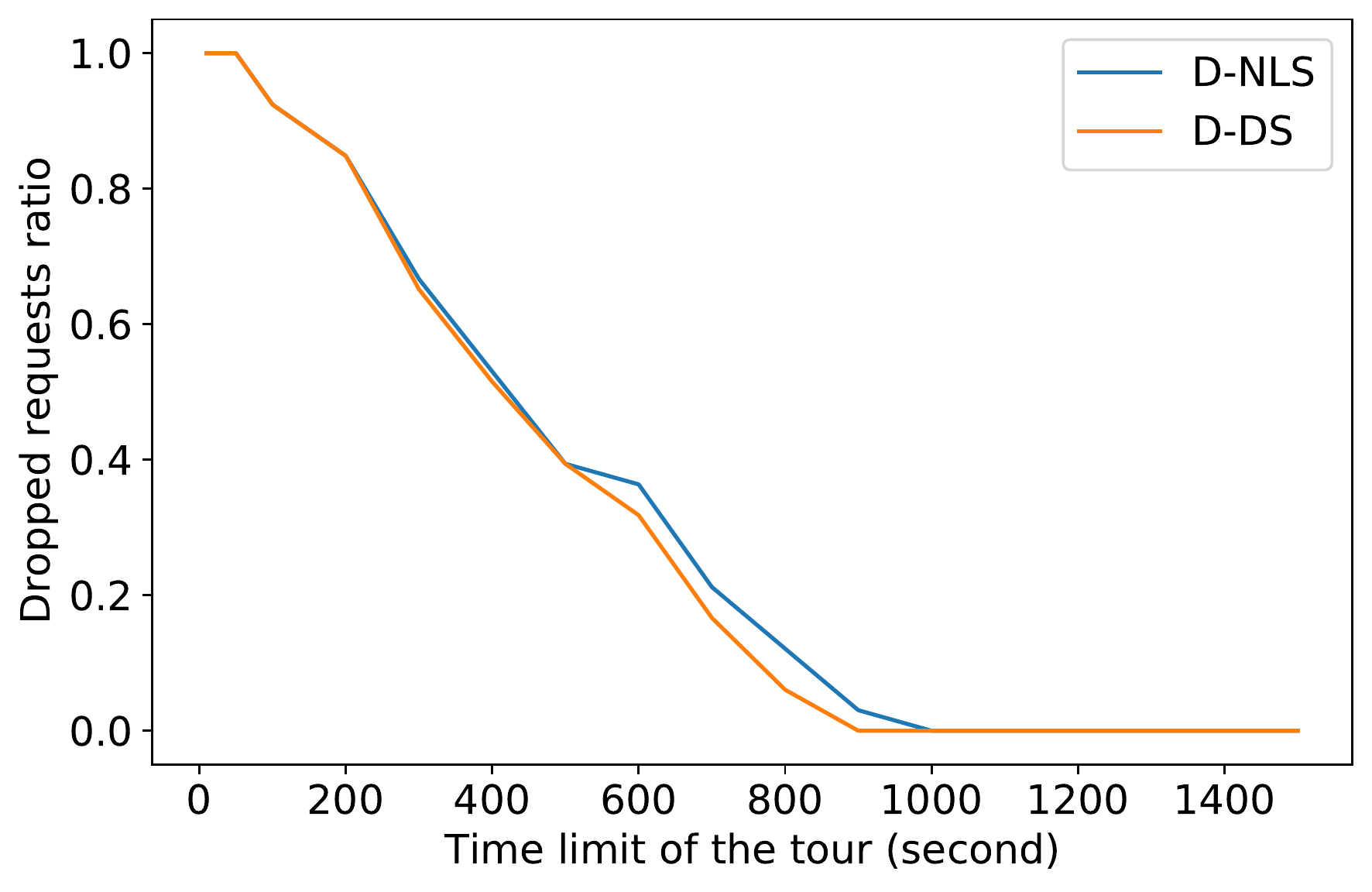}
    \caption{Trade-off between the tour time limit and dropped requests ratio.}
    \label{fig:drop_demand_time_curve}
\end{figure}

With unlimited time, the generated matches and tours can satisfy all human requests. Therefore, the multi-robot planner tries to find the most efficient tours that drop the least number of requests within a limited tour length. We use the smallest test case (\(n_V=4, n_L=10, n_M=10\)) as an example to show the trade-off between the tour time limit and the dropped requests ratio. Fig. \ref{fig:drop_demand_time_curve} shows a smooth and close-to-linear trade-off. Therefore, during a practical situation, we can select an optimal trade-off point accordingly.

\subsubsection{Time Uncertainty and Dropped Requests}\label{sec:computational_time_uncertainty}

We use the smallest test case (\(n_V=4, n_L=10, n_M=10\)) as an example to show the effect of uncertainty in an SMRP. In this parametric study, we assume the traveling time \(T_{kij}\) and visiting time \(T_{ki}\) are Gaussian distributed, and the ground truth standard deviation is 30\% of the expected value, which means the robots' imperfect actions will cause the actual tour time distributed in a range specified by 30\%. To apply the method \modeluncertain{} to consider such uncertainty, the planner also needs to specify an estimated distribution for these time parameters. When the estimated standard distribution (the level of uncertainty) changes, the planner changes its trade-off between satisfying human requests and ensuring the tour time limit. Below, we change the standard deviation of the time distribution input to the planner and show the corresponding trade-off. The metrics used are dropped requests ratio and the average probability of surpassing the time limit during the tour.

\begin{figure}[t!]
    \centering
    \includegraphics[width=1\linewidth]{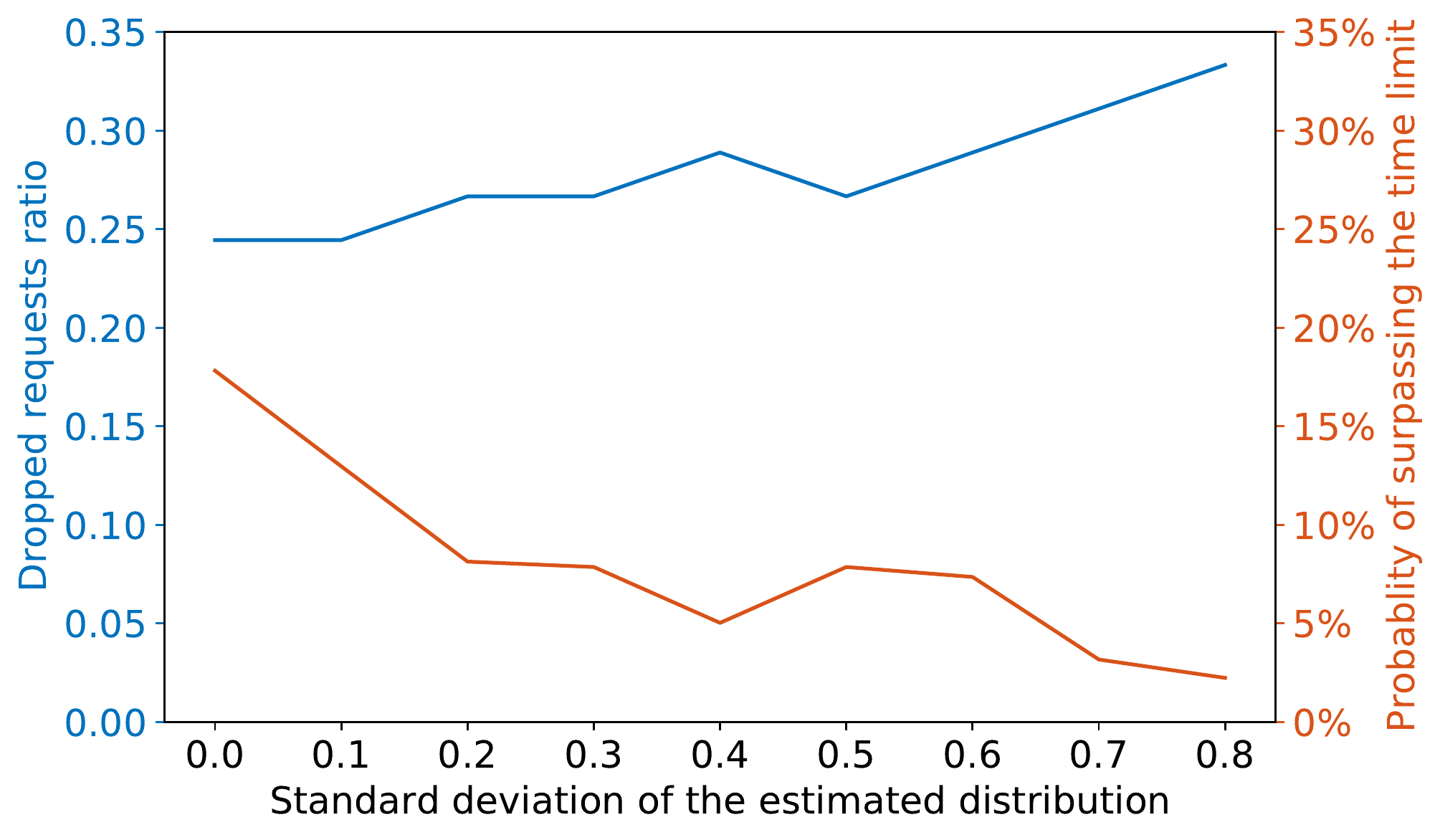}
    \caption{The influence of the estimated time uncertainty. The estimated distribution is the time distribution input to the planner. It does not necessarily equal the ground truth distribution. Here the ground truth distribution has a standard deviation of 0.3.}
    \label{fig:prob_demand_curve}
\end{figure}

According to Fig. \ref{fig:prob_demand_curve} and taking into account the randomness within the optimization, when the estimated standard deviation is larger, more requests are dropped in order to decrease the probability of surpassing the time limit. When a larger uncertainty is estimated, the planner becomes more cautious about time. However, the marginal decrease in the probability becomes small when the estimated standard deviation exceeds a certain threshold. Besides, accurately estimating the uncertainty level, 30\%, provides a good trade-off (both the dropped requests and the probability of surpassing the time limit are small). But a slightly off estimation (e.g., estimate the standard deviation as 20\% or 40\%) still provides a similarly good trade-off. This shows that the optimization is not sensitive to inaccurate uncertainty estimation.

\subsubsection{An Example Case}

Here we use the case in Sec. \ref{sec:computational_time_uncertainty} with 30\% estimated standard deviation to show a plan generated by solving the SMRP using the \modeluncertain{} method.

The 10 people can select their POIs from 10 locations (a small case is used for visualization clarity); there are 4 robots as the tour guides. According to the optimization, the human-robot team number 0-3 contains 2, 3, 2, and 3 humans, respectively (Table \ref{tab:example_team}). The distribution of the POIs and the tours are shown in Fig. \ref{fig:example_route}. Note that POI 9 is dropped by the planner in this example. The plan drops 12 of the 45 requested POIs and satisfies the rest.

\definecolor{colorrobot0}{rgb}{0,0.447000000000000,0.741000000000000}
\definecolor{colorrobot1}{rgb}{0.850000000000000,0.325000000000000,0.0980000000000000}
\definecolor{colorrobot2}{rgb}{0.929000000000000,0.694000000000000,0.125000000000000}
\definecolor{colorrobot3}{rgb}{0.494000000000000,0.184000000000000,0.556000000000000}

\begin{figure}[h]
    \centering
    \includegraphics[width=0.8\linewidth]{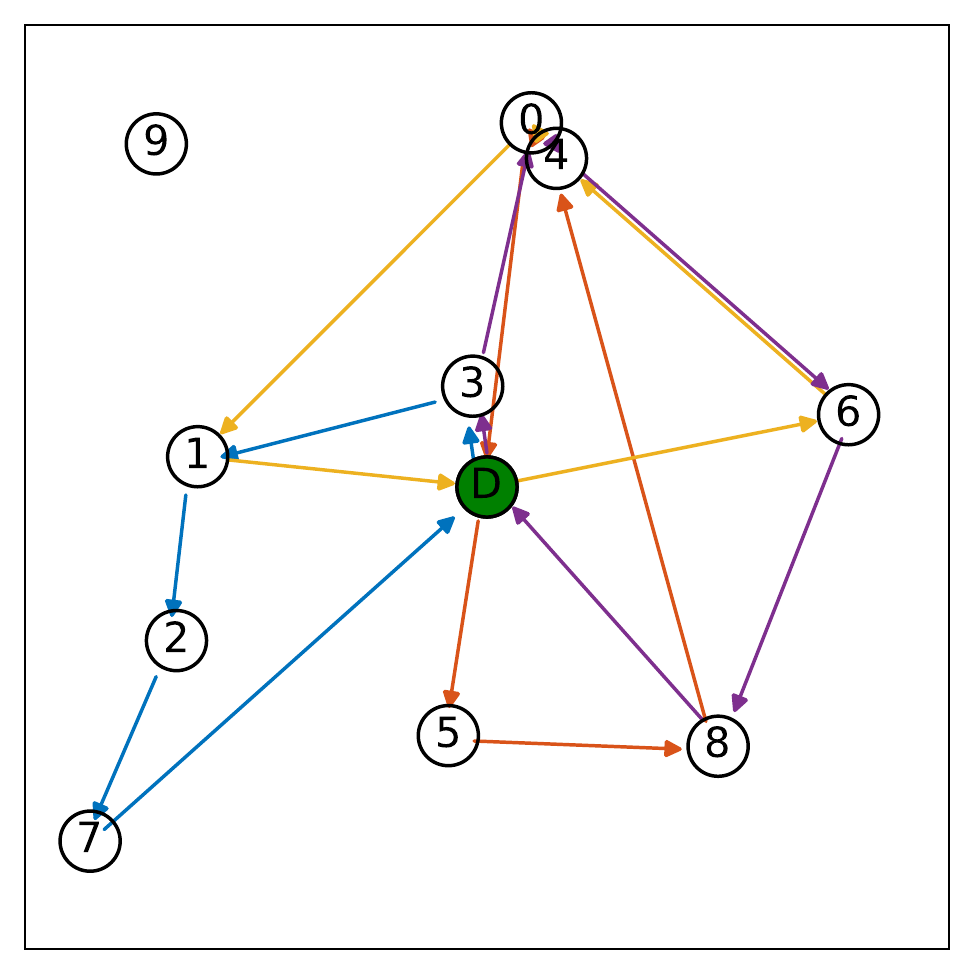}
    \caption{The distribution of the POIs and the routes of the robots. D refers to both the start and terminal. Robot 0: {\color{colorrobot0}blue}, robot 1: {\color{colorrobot1}brick red}, robot 2: {\color{colorrobot2}yellow}, robot 3: {\color{colorrobot3}purple}.}
    \label{fig:example_route}
\end{figure}

\begin{table}[htbp]
  \centering
  \small
  \caption{The planned tour and teams guided by the robots}
    \begin{tabular}{c|lcl}
    \toprule
    Robot & \multicolumn{1}{c}{Tour} & Time (s) & \multicolumn{1}{c}{Humans} \\
    \midrule
    0     & D, 3, 1, 2, 7, D & 465   & 2, 8 \\
    1     & D, 5, 8, 4, 0, D & 517   & 1, 4, 9 \\
    2     & D, 6, 4, 0, 1, D & 522   & 0, 7 \\
    3     & D, 3, 0, 4, 6, 8, D & 426   & 3, 5, 6 \\
    \bottomrule
    \end{tabular}
  \label{tab:example_team}
\end{table}

\subsection{Multi-robot Simulation in Habitat-AI} \label{sec:simulation_experiments}

In this section, we run the whole framework, including the multi-robot planner that solves an SMRP, the local navigation planner, and the simulator.
As described in Sec. \ref{sec:simulation_model}), in the simulator, incorrect actions, non-constant velocity, rotations, as well as the varying visiting time at each POI introduce uncertainties to the traveling and visiting time. Therefore, we first show how different levels of simulated uncertainties will affect the touring time of the human-robot teams.

The SMRP planner assumes estimated distributions of time costs are available.
In reality, we can gather samples to estimate the time uncertainty, but the estimation can be inaccurate.
We did a parametric study on the level of environmental and estimated uncertainties to see how planned tours and actual tour times change accordingly. In the simulation, we constrain the estimated uncertainty to be Gaussian distributions, but our SMRP model with time uncertainty does not assume a specific type of random distribution.

\subsubsection{Varying Environmental Uncertainty}
We fix the plan generated by an SMRP solver with uncertainty (the assumed standard deviation is 40\% of the nominated value), and for the uncertainties in the simulation, we change the rate of correct actions from (100\% to 70\%). Then, we show the results of the actual tour time. Since the result is stochastic, we repeat the test five times for each setup. Note that there are 3 robots, 10 humans, and 22 POI in total for visiting in this simulation. The time limit of the tour is set to 100 seconds.

\begin{figure}[t]
    \centering
    \includegraphics[width=0.9\linewidth]{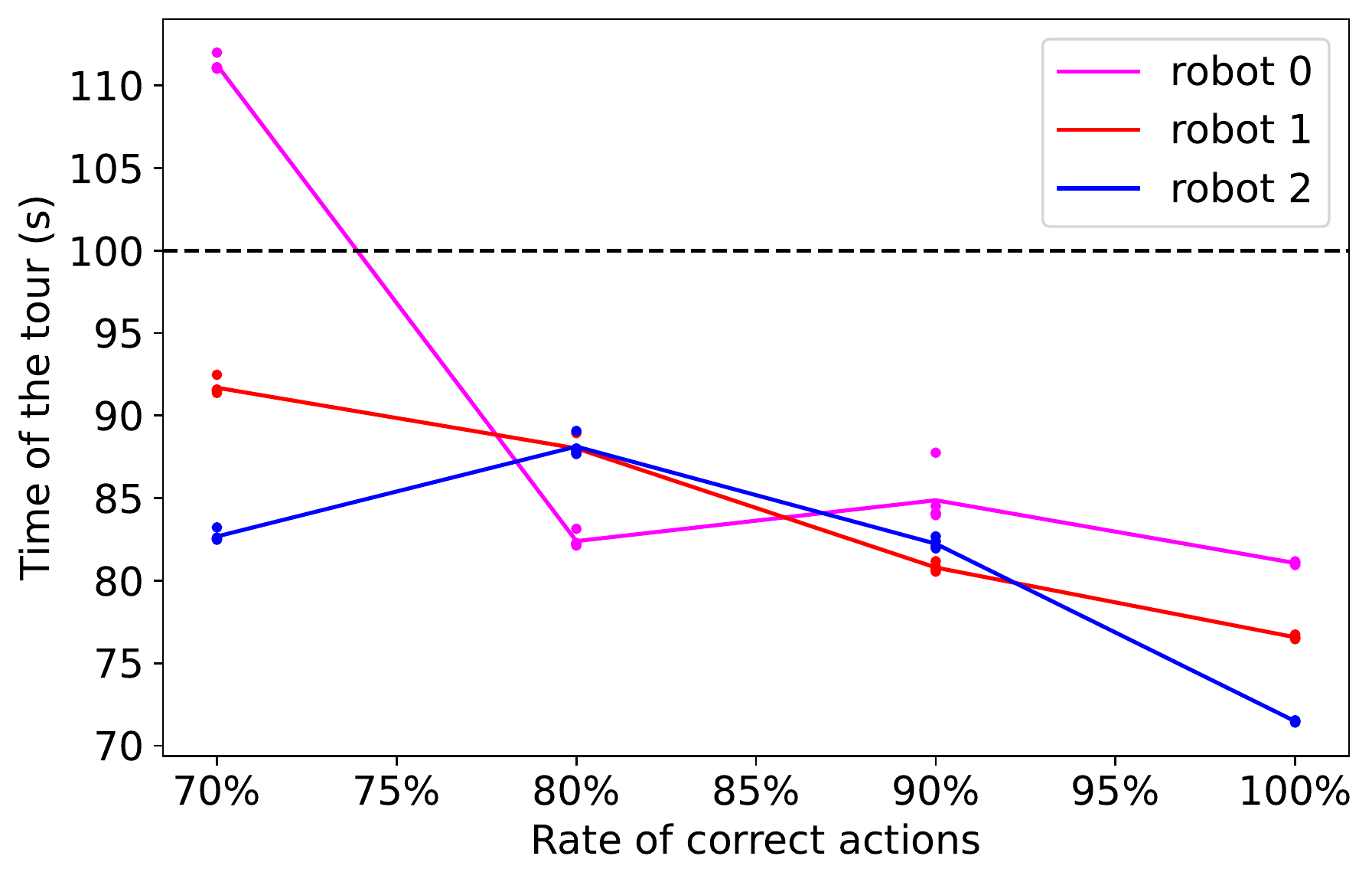}
    \caption{The actual tour time of the three robots when the environmental uncertainty changes.
    The dots show the results of multiple random trials, while the line shows the mean.
    The expected tour times according to the planner are 83, 82, and 75 seconds, respectively.}
    \label{fig:sim_rate}
\end{figure}

\begin{figure}[t]
    \centering
    \includegraphics[width=1.0\linewidth]{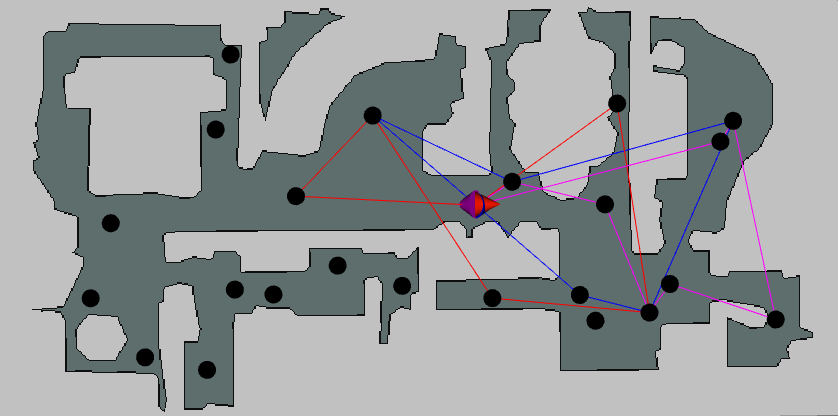}
    \caption{The planned tour of the three robots assuming the standard deviation is 40\%. Between two POIs, straight lines instead of the actual trajectories are shown for clarity.}
    \label{fig:sigma4}
\end{figure}

According to Fig. \ref{fig:sim_rate}, with the touring routes fixed (Fig. \ref{fig:sigma4}), the tour time generally increases when the rate of correct actions decreases. Apart from that, for the situations with \(\geq\) 80\% accuracy, the generated plans are always completed within the time limit. 

We note that the time uncertainty is assumed as Gaussian distribution, but the actual time uncertainty introduced by incorrect actions may not necessarily be Gaussian. In addition, the standard deviation of the actual time distribution might not be 40\%. Nevertheless, a fixed and possibly inaccurate uncertainty estimation (40\%) can ensure the time constraints of the tour under different correct rates (80\% to 100\%).
This shows that the generated plan has a relatively wide margin with respect to the uncertainties during the tasks. Moreover, the robust planning framework does not require a perfectly accurate uncertainty estimation to work.

\subsubsection{Varying Estimated Uncertainty}

\begin{figure}[t]
    \centering
    \includegraphics[width=0.9\linewidth]{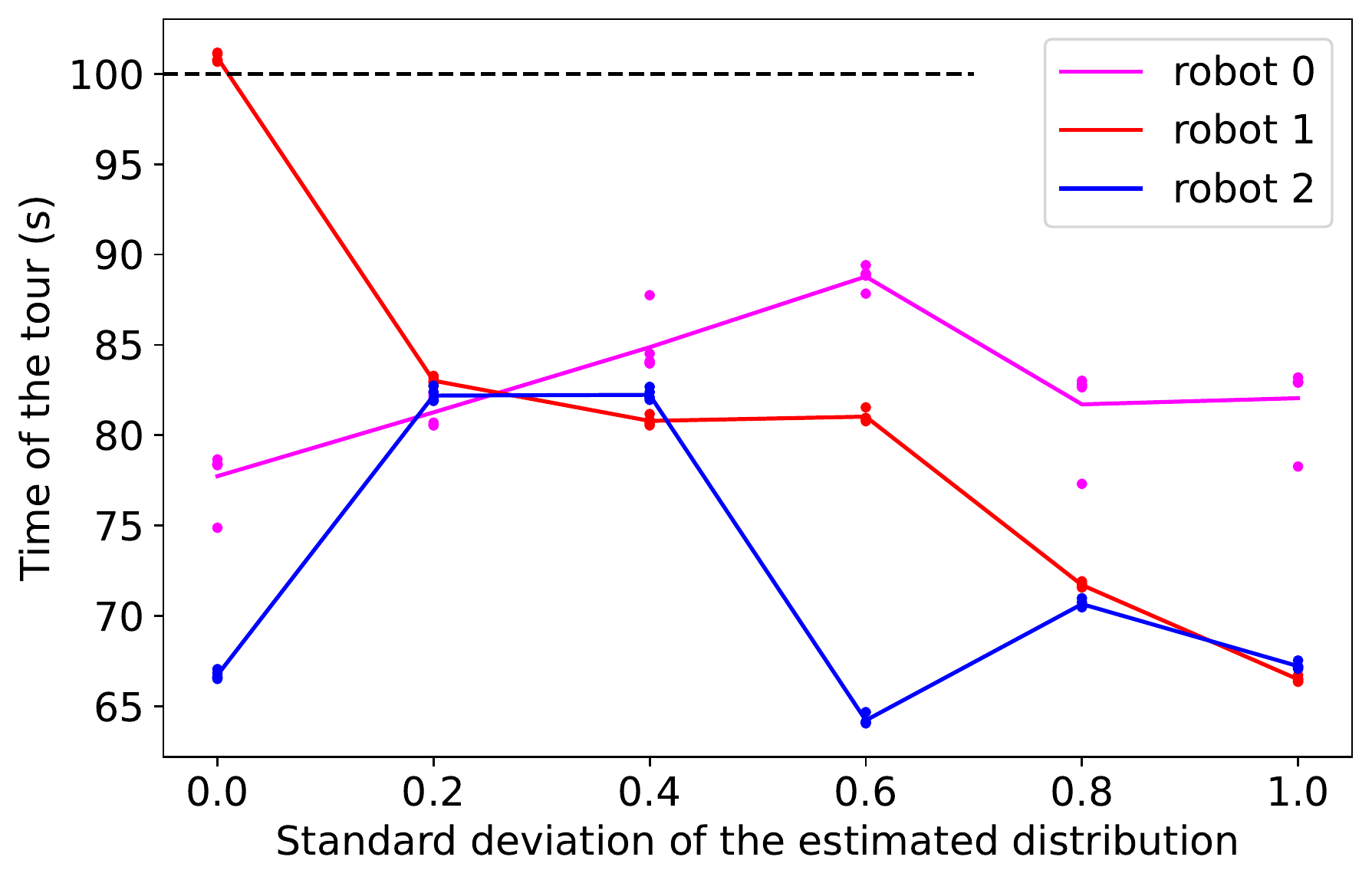}
    \caption{The actual tour time of the three robots when the estimated uncertainty changes. The dots show the results of multiple random trials, while the line shows the mean.}
    \label{fig:sim_sigma}
\end{figure}

In this subsection, we fix the rate of correct actions in the simulation and vary the estimated uncertainty of the planner. We input Gaussian distributions to the planner and set the uncertainty for traveling between two POIs a fraction (0\% to 100\%) of the nominal value (proportional to traveling distance). Again, under one setup, the tests are repeated five times. The actual tour times are shown in Fig. \ref{fig:sim_sigma}. According to the figure, there is no clear trend when the standard deviation of the estimated distribution increases. For this specific case, all the plans that consider the uncertainties end up not surpassing the time limit. Again, this is a sign that a perfect estimation of the uncertainty parameters is not a hard requirement for the multi-robot SMRP planner model with uncertainty.

\begin{figure}[t]
    \centering
    \includegraphics[width=1.0\linewidth]{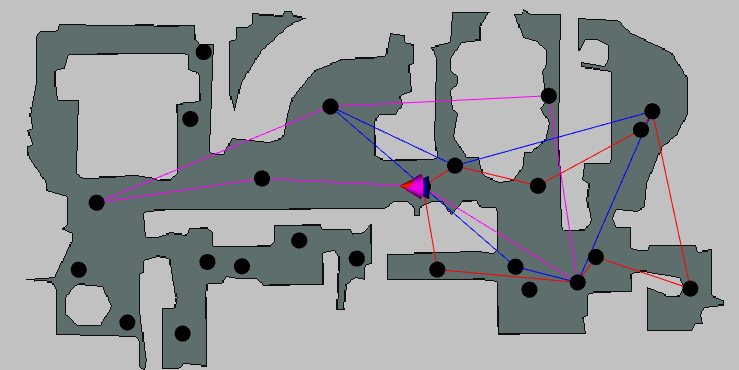}
    \caption{The planned tour of the three robots assuming no uncertainty (standard deviation is 0\%).}
    \label{fig:sigma0}
\end{figure}

Finally, the planned tour with 0\% and 40\% standard deviations are shown in Fig. \ref{fig:sigma0} and Fig. \ref{fig:sigma4}, respectively. And it can be seen that the tours in Fig. \ref{fig:sigma0} are indeed longer and, therefore, more aggressive. Here, a more aggressive tour tries to address more human requests but is more likely to surpass the tour time limit.

\section{Conclusions and Future Work}\label{sec:conclusion}

This paper presents a multi-robot framework that deals with robotic tour guiding in a partially known environment with uncertain traveling and visiting times.
A two-level structure is described in the paper: a centralized multi-robot planner (the focus of the paper) generates the tours of the robots and assigns the humans to robots to maximize the number of satisfied human requests (places of interest); a distributed local navigation planner decides the actual trajectories for a single robot according to the global tour routing plan.
The problem considered in the multi-robot planner is generalized and modeled as a simultaneous matching and routing problem, which simultaneously optimizes the human-robot matching and robot routing problem. The uncertainties in the traveling and visiting times are considered, and robust plans are generated by penalizing the expected tour times that pass a predefined threshold. The mixed-integer bilinear program that represents the above planning problem is solved through both an exact branch and cut algorithm as well as an innovative large neighborhood search method. 
The algorithms are evaluated through comprehensive computational investigations.
Then, the robustness of the whole framework is evaluated in a photo-realistic simulation that captures multiple practical uncertainties.

Results demonstrate that, through the large neighborhood search, the planner is scalable to the number of robots, humans, and POIs and can output plans with low dropped requests. The largest case tested involves 50 robots, 250 humans, and 50 POIs. Second, results show that the planner makes a smooth trade-off under the time limit for the tours, which provides valuable information for tour design. Finally, parametric study and simulation evaluation demonstrate that the planner can generate robust tours that keep the time constraints in practice with crude uncertainty estimations.

Future work will consider dynamic replanning mechanisms to address real-time disturbances or failures, as well as a more consolidated local planner that can deal with local disturbances. These are options to further improve the robustness of the tour guiding. Heterogeneity within the robots will be considered too. Examples include different sensors and actuators that allow different robots to perform distinct gestures and movements during the guidance. Real-world experiments will be a further step to demonstrate the practicality of our proposed system.

\begin{acknowledgment}
Distribution A. Approved for public release; distribution unlimited (OPSEC 5767).
This research has been partially supported by the Automotive Research Center, a US Army center of excellence for modeling and simulation of ground vehicles.

\end{acknowledgment}

{\small
\bibliographystyle{asmems4}
\bibliography{bib/ieee_full, bib/strings_full, ms}
}

\end{document}